# Development of a Neural Network-based Method for Improved Imputation of Missing Values in Time Series Data by Repurposing DataWig

**Daniel F. Zhang[1,2]**

[1] Hackley School, 293 Benedict Ave, Tarrytown, NY
[2] Current Address: Department of Computer Science, Rice University, 6100 Main St, Houston, TX
Email: danielfzhang@gmail.com

## Abstract

Time series data are observations collected over time intervals. Successful analysis of time series data captures patterns such as trends, cyclicity and irregularity, which are crucial for decision making in research, business, and governance. However, missing values in time series data occur often and present obstacles to successful analysis, thus they need to be filled with alternative values, a process called imputation. Although various approaches have been attempted for robust imputation of time series data, even the most advanced methods still face challenges including limited scalability, poor capacity to handle heterogeneous data types and inflexibility due to requiring strong assumptions of data missing mechanisms. Moreover, the imputation accuracy of these methods still has room for improvement. In this study, I developed tsDataWig (time-series DataWig) by modifying DataWig, a neural network-based method that possesses the capacity to process large datasets and heterogeneous data types but was designed for non-time series data imputation. Unlike the original DataWig, tsDataWig can directly handle values of time variables and impute missing values in complex time series datasets. Using one simulated and three different complex real-world time series datasets, I demonstrated that tsDataWig outperforms the original DataWig and the current state-of-the-art methods for time series data imputation and potentially has broad application due to not requiring strong assumptions of data missing mechanisms. This study provides a valuable solution for robustly imputing missing values in challenging time series datasets, which often contain millions of samples, high dimensional variables, and heterogeneous data types.

## Introduction

Time series data are a collection of observations obtained through repeated measurements over time. Time series data are everywhere. Examples include weather records, economic indicators and patient health evolution metrics. Analysis of time series data has been used to reveal major patterns such as trends, seasonality, cyclicity and irregularity, which often lead to informed decisions that are crucial for research, healthcare, business and policy making (*1*).

A successful analysis of time series data depends on data reliability and completeness of collected information. However, missing observations can commonly occur during numerous processes involved in data collection, for example communication errors, failure of data-generating sensors or power outages (*2*). Missing data present significant challenges to data analysis and may lead to undesirable outcomes including inaccurate predictions or poor decisions (*2*). In addition, time series data are sequential, an attribute that differs time series data from other data types. Hence, methods to replace missing data with appropriate alternative values for time series data type are needed.

In the past decades, various approaches have been developed to address missing values in time series data (*3-5*). A simple solution is to omit missing data, but this is very risky if the missing percentage of the data is big enough to disrupt the result of the analysis. Another solution is to fill in the missing values with substituted values, a process called data imputation. Mean imputation, a method to replace missing values with the mean of the available values in the dataset, comes as a popular solution due to its simplicity. However, this method can lead to severely biased outcomes caused by reduced variability of the data (*6*). Although more advanced statistics modeling and machine learning approaches that are more advanced and specific to time series data have been proposed to better impute missing data, these methods often do not capture variable correlations and/or require strong assumptions of the underlying data generating process, which limit their applications (*7-9*).

Recently, a few neural network-based methods, such as Recurrent Neural Networks (RNN) and Gated Recurrent Unit (GRU), have been shown to achieve state-of-the-art results due to their abilities to capture the patterns of missingness in time series data (*10, 11*). However, scalability and capability to handle heterogeneous data types including free form text fields remain challenging to these methods. Moreover, the accuracy of such methods still has much room to improve (*11*). DataWig, a deep neural network developed for imputing missing data, has been demonstrated to be scalable and capable of handling diverse data, but its use in processing time series data types has not been documented (*12*). The goal of this study is to modify DataWig to handle time series data, and to benchmark it against currently most widely used and most advanced approaches for the task of time series data imputation.

## Materials and Methods

**Equipment**

**Computer**: Dell XPS 8940 Tower, model #XPS8940-7354BLK-PUS, 10th Gen Intel® Core™ i7-10700 Processor 2.9GHz, 64 GB memory (RAM), 2TB 7200RPM SATA Hard Drive and 1TB PCIe NVMe M.2 Solid State Drive, NVIDIA® GeForce GTX™ 1660 Ti

**Operating system**: Ubuntu 20.04

**Computational packages**

**DataWig and dependencies:**

DataWig, a neural network-based Python package, version 0.2.0 (*12*);

MXNet, a deep learning Python framework, version 1.7.0 (*13*);

Scikit-learn, a machine-learning Python library, which DataWig uses for feature extraction, scaling and evaluation metrics for hyperparameter optimization, version 0.23.2 (*14*);

Pandas, a Python package for data structure conversion to input data to DataWig (DataWig is designed to take pandas DataFrame as input, because pandas allows the user to easily read data from a .csv file, and turn it into a data structure called a pandas DataFrame, which is usable for machine learning), version 1.1.4.

**TimeSynth:**

A Python package that simulates time series datasets using various models, version 0.2.4 (http://github.com/TimeSynth/TimeSynth)

**Packages for conventional imputations:**

Scikit-learn, a machine-learning Python library that has several conventional methods, version 0.23.2;

Fancyimpute, a Python library for imputation with simple fill methods and other conventional methods, version 0.5.5.

**Data masking by different data missing mechanisms**

**Missing Completely at Random (MCAR):**

There is no relationship between the missingness of the data and any values (observed or missing). Those missing data points are a random subset of the data. There is nothing systematic going on that makes some data more likely to be missing than others (*2*).

**Missing at Random** (**MAR**):

There is a systematic relationship between the propensity of missing values and the observed data, but not the missing data (*2*).

**Missing Not at Random** (**MNAR**):

There is a relationship between the propensity of a value to be missing and its own values (*2*).

**Real-world datasets**

**Air quality dataset:**

Consists of PM2.5 measurements from 36 monitoring stations in Beijing. The measurements are collected hourly from 2014/05/01 to 2015/04/30. Overall, 13.3% of the values are missing. The raw data can be downloaded from: https://archive.ics.uci.edu/ml/datasets/Beijing+PM2.5+Data.

**Health-care dataset:**

Consists of 4000 multivariate clinical time series from intensive care units (ICU). Each time series contains up to 42 measurements such as albumin concentration, heart-rate etc., which are irregularly sampled at the first 48 hours after the patient's admission to ICU. This is a sparse dataset with up to 75% of missing values. The raw data can be downloaded from: https://physionet.org/content/challenge-2012/1.0.0/.

**Human activity dataset:**

Contains records of five people performing 11 different activities such as walking, falling, sitting down, etc. Each person wore four sensors on the person's left/right ankle, chest, and belt. Each sensor recorded 3-dimensional coordinates for about 20 to 40 milliseconds. The raw data can be downloaded from: https://archive.ics.uci.edu/ml/datasets/Localization+Data+for+Person+Activity.

**Encoding the values for the time variable in time series data**

Two strategies are used for encoding the values for the time variable in the three real-world datasets.

1. Encoding cyclical time values: Cyclical time values are in such format as "yy/mm/dd/hh" (used in air quality dataset) and "hr/min" (used in health-care dataset). The datetime variable "yy/mm/dd/hh" is cut into four variables (year, month, day and hours), and each of these variables are then decomposed into two more variables (except for year, which was directly converted to an integer) by creating a sine and a cosine facet, which retains the cyclical nature of these variables; for example, hour 24 is closer to hour 0 than to hour 21, and month 12 is closer to month 1 than to month 10, etc. The following equations are used for the data encoding:

   $mSine = sin\left(\frac{2\pi m}{12}\right); mCosine = cos\left(\frac{2\pi m}{12}\right)$, where $m$ is an integer which represents a month.

   $dSine = sin\left(\frac{2\pi d}{30}\right); dCosine = cos\left(\frac{2\pi d}{30}\right)$, where $d$ is an integer which represents the date within a month.

   $hSine = sin\left(\frac{2\pi h}{24}\right); hCosine = cos\left(\frac{2\pi h}{24}\right)$, where $h$ is an integer which represents the hour in a 24-hour format.

   The time variable in "hr/min" format is converted to minutes in integer form with the following equation: $totalmin = hr * 60 + min$. The minute data is then sine/cosine transformed with the following equations:

   $minSine = sin\left(\frac{2\pi * min}{t}\right); minCosine = cos\left(\frac{2\pi * min}{t}\right)$, where $t$ is the total number of minutes in a day ($t = 24 * 60$).

2. Encoding continuous time values: Continuous time series data are data collected from continuous measurements during certain periods of time with no clear cyclical pattern, such as the human activity data (measuring walking and sitting, etc. for 30 to 60 minutes). This type of time value is converted into integers using the smallest time unit. For example, 03:25:127 (minute:second:millisecond) is encoded as 205127 (3*60*1000+25*1000+127).

**Data analysis**

**Evaluations on imputed values**:

The correlation between actual values (ground-truth) vs. imputed values by DataWig in simulated datasets and real-world datasets is calculated to evaluate the validity and performance of DataWig in imputing time series data. Resulting MRE (mean relative error)

6values from imputation for real-world datasets by DataWig are calculated to benchmark the imputation accuracy of DataWig.

**Calculation of Pearson R Correlation (r):**

$$r_{xy} = \frac{\sum_{i=1}^{n}(x_i - \bar{x})(y_i - \bar{y})}{\sqrt{\sum_{i=1}^{n}(x_i - \bar{x})^2}\sqrt{\sum_{i=1}^{n}(y_i - \bar{y})^2}}$$

Where, $x_i$ is the $i^{th}$ imputed value, $\bar{x}$ is the average of all imputed values, $y_i$ is the $i^{th}$ ground-truth value and $\bar{y}$ is the average of all ground-truth values.

**Calculation of Mean Relative Error (MRE):**

$$MRE = \frac{\sum_{i=1}^{n}|x_i - y_i|}{\sum_{i=1}^{n}|y_i|}$$

Where $x_i$ is the $i^{th}$ imputed value and $y_i$ is the $i^{th}$ ground-truth value.

**Calculation of Mean Square Error (MSE):**

$$MSE = \frac{1}{n}\sum_{i=1}^{n}(x_i - y_i)^2$$

Where $x_i$ is the $i^{th}$ imputed value and $y_i$ is the $i^{th}$ ground-truth value.

**Calculation of 2-Norm of vector:**

$$\|A\|_2 = \sqrt{\sum_{i=1}^{n}(x_i - y_i)^2}$$

Where $x_i$ is the $i^{th}$ imputed value and $y_i$ is the $i^{th}$ ground-truth value.

# Results

## Summary of representative methods for imputing missing values in time series data

Literature search was performed to identify existing methods mainly in the field of time series data imputation. These methods can be classified into three categories, which are simple imputation, classic imputation and neural network approach, based on both the development history of data imputation methods and the level of technical difficulty involved with each method. The representative methods for each category are selected and presented in Table 1.

**Table 1. Summary of methods for imputing missing values in time series data**

| Category | Method | Function | Comments | Reference |
|---|---|---|---|---|
| Simple imputation | Mean | Replace the missing values with corresponding global mean | Easy and simple | D. Little et al., Wiley, New York., P.381, 1987; I.B. |
| | | | Does not preserve the relationships among variables | |
| Classic imputation | KNN (K-Nearest Neighbor) | Use k-nearest neighbor to find the similar samples and imputing the missing values with wighted average of its neighbors | Imputates via classification | J. Friedman et al., volume 1. Springer series in statistics Springer, Berlin, 2001. |
| | | | No training required | |
| | | | Does not work well with large and high demensional datasets | |
| | MF (Matrix Factorization) | Factorize the data matrix into two low-rank matrices and then try to reconstruct the original matrix | Imputates via classification | J. Friedman et al., volume 1. Springer series in statistics Springer, Berlin, 2001. |
| | | | Data driven and can handle high demensional data with many missing values | |
| | MICE (Multiple Imputation by Chained Equations) | Create multiple imputations with chained equations to fill the missing values | Popular method, does not suffer from bias values like single imputation | S. Buuren et al., J Statist Softw. 2010 |
| | | | Assumes that data are MAR (missing at random) | |
| | ImputeTS | Impute missing value using the state space model and kalman smoothing | Widely used package | S. Moritz et al, The R Journal |
| | | | For univariate time series | |
| | STMVL (Spatio-Temporal MultiView-based Learning) | STMVL is the state-of-the-art method for air quality data imputation by further utilizing the geo-locations | Tailored for imputing the missing values in air quality time series data | X. Yi et al., IJCAI, 2016 |
| Neural network approach | GRU-D (Gated Recurrent Unit) | Impute each missing value by the weighted combination of its last observation and the global mean, together with a recurrent | Based on recurrent neural network | Z. Che et al., Scientific reports, 2018 |
| | | | Tailored for handling missing values in health-care data | |
| | M-RNN (Multi-directional Recurrent Neural Networks) | Impute the missing values according to hidden states in both directions in RNN | Treats the imputed values as constants | J. Yoon et al., arXiv:1711.08742v1 |
| | | | Does not consider the correlations among different missing values | |
| | BRITS | Impute the missing values based on bi-directional recurrent neural networks without any specific assumption | Handles multiple correlated missing values in time series | Bießmann F. et al., Journal of Machine Learning Research, 2019 |
| | | | Generalizes to time series with nonlinear dynamics underlying | |
| | | | Provides a data-driven imputation procedure and applies to general settings with missing data | |

The listed methods in each category are either the most commonly used, the most advanced and/or specialized to process certain types of time series data such as air quality and health care data, and therefore have been considered as the gold standard method for imputation of such data types. These representative methods can be used as benchmarks for testing candidate methods for imputing missing values in time series data.

**Building a computational pipeline to repurpose and benchmark DataWig for imputing time series data**

Through literature review, DataWig, a neural network-based Python package for imputing missing values, is the only method I identified that can be both scalable and capable of handling heterogeneous datasets (*12*). However, the application of DataWig in time series data imputation has never been documented. To evaluate the potential of applying DataWig to imputing missing values in time series data, I built a computational pipeline for validating the performance of DataWig.

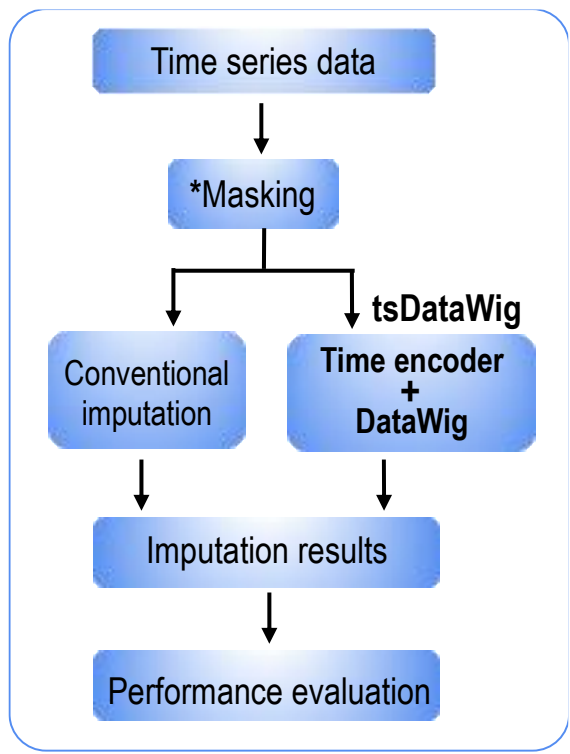

**Figure 1. Computational pipeline for imputing time series data.**

As illustrated in Fig. 1, I installed the DataWig Python package, packages for conventional data imputing methods and TimeSynth, and a package for simulating time series data, in a Linux Operating System. In order to streamline the testing, I wrote programs in Python to generate simulated data from TimeSynth, convert the data format to be compatible with DataWig's input format, mask the simulated data to create missing values, run DataWig to impute the missing values, save the final results and perform data analysis. To compare DataWig with the representative methods for filling missing values in time series data, I selected and installed packages for the representative conventional imputing methods, tested and optimized the versions of dependencies on DataWig to ensure their compatibility in the same computational environment. I wrote a Python script as an add-on to DataWig package for encoding time values, as described in the Materials and Methods section, to repurpose DataWig for time series data imputation. I also re-wrote some of the source code in the conventional method packages for them to accept newer dependencies, as several of the dependencies of DataWig were not compatible with the newest versions of the conventional imputation packages. For data analysis as part of the evaluation metrics, I wrote Python scripts to implement the methods for evaluating the error of the imputed values, which includes Pearson R correlation, mean relative error (MRE), mean squared error (MSE) and 2-Norm of a vector, and to finally plot the results for visualization at the end of each run.

**Validating repurposed DataWig on imputation of missing values using synthetic time series datasets**

DataWig was first tested on synthetic time series datasets using the built computational pipeline illustrated in Figure 1. The synthetic datasets were generated by a pseudo periodic signal model from the TimeSynth python package. The simulated dataset has 30 samples, and each sample has 100 regularly spaced time points from 0 to 20, as shown in Figure 2A. The missing data points were achieved by masking 30% of the values in the dataset, using each of the three data missing mechanisms: Missing Completely At Random (MCAR), Missing At Random (MAR) and Missing Not At Random (MNAR) (*2*). Each masked dataset was formatted into a 100 by 31 Pandas DataFrame, and the time values in the simulated dataset were ordinally encoded as the input to DataWig for imputing missing data.

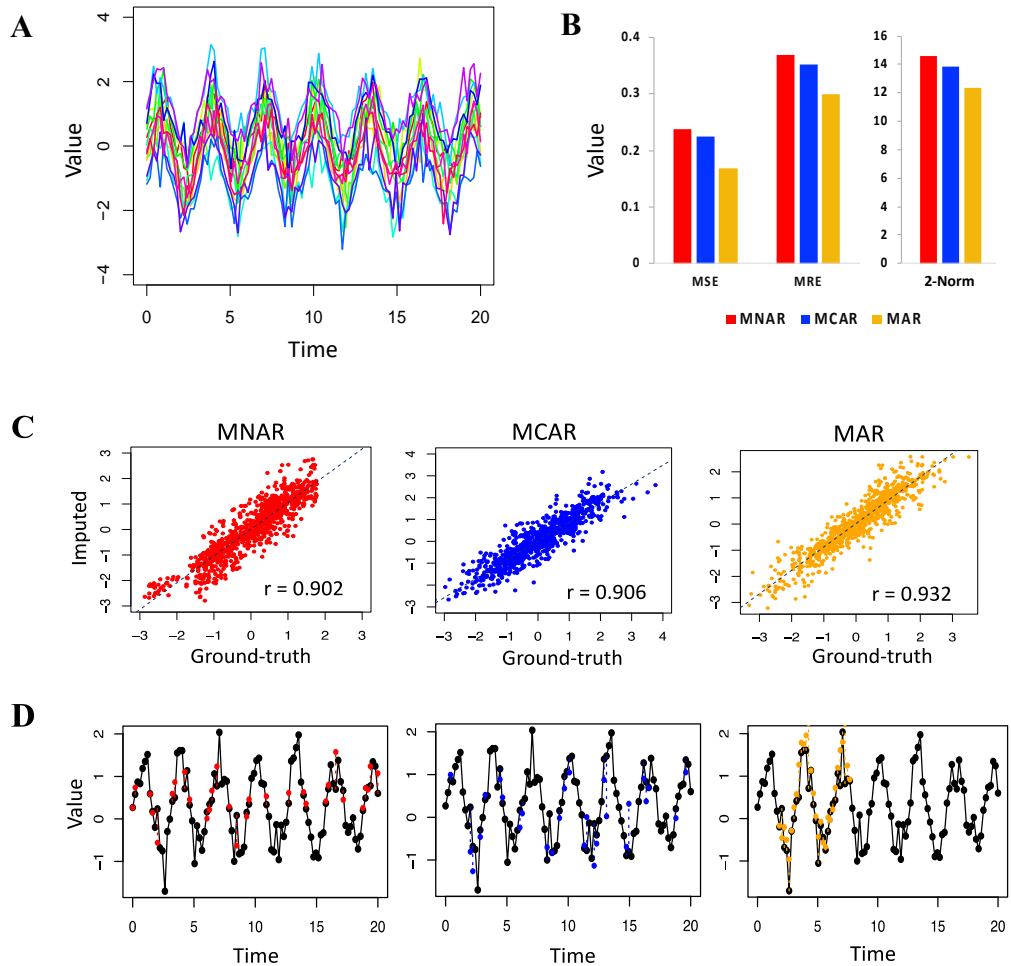

**Figure 2. Evaluation of DataWig performance using simulated data.** (A) The simulated dataset generated by a pseudo periodic signal model from the TimeSynth package. Each line represents a simulated time series data sample. (B) Evaluation of DataWig performance on the simulated data masked by different missing mechanisms (MNAR, MCAR and MAR). MSE: mean squared error; MRE: mean relative error; 2-Norm: 2-norm of vector. (C) Correlations between imputed values and the ground-truth values in simulated datasets. (D) A case example of comparison between simulated values (ground-truth) and values masked and then imputed by DataWig. All data points are from one simulated sample. Black dots are original data, and red, blue and orange dots represent imputed values for masked data points using missing mechanisms of MNAR, MCAR and MAR, respectively.

The results from imputing the simulated data demonstrate that DataWig performs almost equally well on the datasets masked with three different missing mechanisms, with slight favor to missing type MAR, as shown in Figure 2B, which plots MSE (mean square error), MRE (mean relative error) and 2-Norm of vector, the standard measurements for comparing model performance. This suggests that DataWig can be used without strong assumption of data missing mechanisms. In addition, DataWig imputed values correlates well with the ground-truth values (defined as the values prior to being masked in the simulated dataset) as shown in Figure 2C, confirming that DataWig is able to fill the missing values in time series data with good accuracy. Consistent with the good correlations, the imputed and ground-truth data points are generally close to each other, as illustrated in Figure 2D, which shows difference/similarity of imputed and ground-truth values at a single data point level.

Taken together, these results suggest that DataWig, a package that has demonstrated capability of handling heterogeneous data and scalability in non-time series data, can be repurposed to impute time series data, despite previously only having been tested on cross-sectional data (*12, 14*).

**Benchmarking repurposed DataWig (tsDataWig) using real-world datasets**

In one of the studies I reviewed, BRITS, a model based on bi-directional recurrent neural network developed to impute missing values without any specific assumption (*11*), was compared to many other methods, including the representative methods described above, using real-world air quality, healthcare and localization for human activity datasets (detailed in Materials and Methods) with additional 10% data missing at completely random.

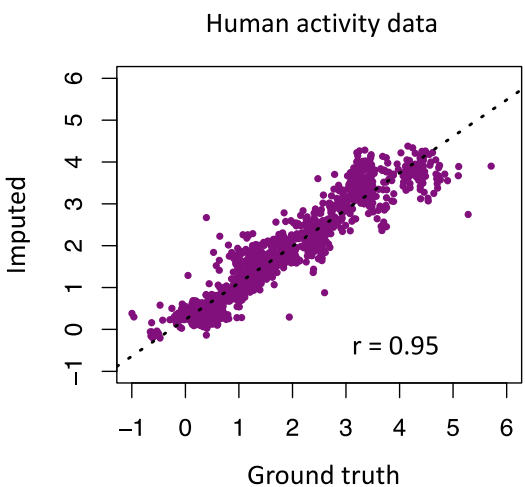

**Fig. 3.** Imputed values by tsDataWig show good correlation with ground truth values. The plot shows the representative result using the real-world human activity dataset.

DataWig has been developed with the capacity to scale to datasets with millions of rows and handle tables with heterogeneous data types, such as numerical, categorical and unstructured text data. Furthermore, it has also been shown that DataWig outperforms state-of-the-art methods for filling missing values in non-time series data (*12*). For benchmarking DataWig on imputing time series data, I focused on the comparison of imputation accuracy between DataWig and BRITS, using the same public datasets as the ones being used in the BRITS study and with the same additional 10% data missing at completely random applied. Two functions for encoding time values were created as described in the Materials and Methods section to enable DataWig to appropriately handle time series data. Overall, the imputed values by repurposed DataWig, denoted as tsDataWig (time series DataWig), show a good correlation with ground truth values in these datasets, as shown in Figure 3.

**Table 2. Imputing performance comparison (Mean Relative Error)**

| Method | Air Quality | Healthcare | Human Acitivity |
|---|---|---|---|
| Mean | 0.7797 | 0.6561 | 0.9643 |
| KNN | 0.4185 | 0.5215 | 0.5854 |
| MF | 0.3925 | 0.6797 | 1.1044 |
| MICE | 0.3852 | 0.7250 | 0.5794 |
| ImputeTS | 0.2751 | 0.5420 | 0.4565 |
| STMVL | 0.1740 | N/A | N/A |
| GRU-D | N/A | 0.7758 | 0.7005 |
| M-RNN | 0.2016 | 0.6187 | 0.3119 |
| BRITS | 0.1665 | 0.3872 | 0.2759 |
| **tsDataWig** | **0.1646** | **0.2011** | **0.1612** |

To evaluate the performance of tsDataWig, MRE (mean relative error, one of the standard methods for evaluating model performance) was calculated using the imputation results generated from the tsDataWig pipeline. The MREs from the imputations of the same datasets by some of the existing representative methods were retrieved from the BRITS benchmark study (*11*) to compare with the performance of tsDataWig. The comparison results are shown in Table 2. Notably, simply applying naïve mean imputation is very inaccurate. Although the conventional imputation methods such as MF (*15*) and STMVL (*16*) are able to perform better than mean imputation, their performance is unstable and varies in different tasks, demonstrating the limits of such methods. For example, the MF algorithm shows good accuracy on the air quality data, but very poor accuracy on the human activity data. While STMVL performs well on the air quality data, it cannot be applied to other data types. The performances of the neural network-based methods suggest more accurate imputation results can be achieve using appropriate neural network frameworks, and tsDataWig has the most favorable performance among all the neural network-based methods (GRU-D, M-RNN, BRITS and tsDataWig).

**Conclusion and discussion**

This study successfully identified and repurposed DataWig (tsDataWig) to appropriately handle time series data. Using real-world datasets, tsDataWig is validated as a method for imputing missing values in time series data with higher accuracy than existing time series data imputation methods including other neural network-based models (e.g., BRITS). In addition, the results of this study demonstrated that tsDataWig is very much capable of handling missing data of all known missing mechanisms (MCAR, MAR and MNAR), and therefore can potentially be applied broadly with no need to make strong assumptions on data missing mechanisms. Furthermore, this study validated the capacity of tsDataWig handling high dimensional variables in time series data using three complex real-world datasets.

Due to the limit of current computational equipment, the scalability of tsDataWig has been tested only up to tens of thousands of variables instead of DataWig's capacity of millions of variables, and the performance of tsDataWig on heterogeneous data types was only evaluated on numeric and categorical datasets but not on free form text datasets. In addition, I observed that tsDataWig maintained its good performance with up to 70% missing values after a wide range of percentages of missing values having been tested (data not shown). Future direction of this project includes further validation of tsDataWig using more computing power (HPC/GPU), and development of a new imputation method based on tsDataWig to further improve performance, such as better handling data with over 70% missing values.

Imputing missing values is still very challenging in certain time series data such as the combination of longitudinal gene expression profiles of patient blood samples and clinical meta-information, which often contain very high denominational variables, large samples and heterogeneous data types. The success of this study may provide valuable solutions to address these challenges.

**Code Availability**

Code and implementation can be found on GitHub: [https://github.com/danielzhang-hackley/tsDataWig](https://github.com/danielzhang-hackley/tsDataWig).

**Reference**

1. T.-c. Fu, A review on time series data mining. *Engineering Applications of Artificial Intelligence* **24**, 164-181 (2011).
2. I. Pratama, A. Permanasari, I. Ardiyanto, R. Indrayani, *A review of missing values handling methods on time-series data*.  (2016), pp. 1-6.
3. S. Moritz, A. Sardá-Espinosa, T. Bartz-Beielstein, M. Zaefferer, J. Stork, Comparison of different Methods for Univariate Time Series Imputation in R.  (2015).
4. M. Spratt *et al.*, Strategies for Multiple Imputation in Longitudinal Studies. *American Journal of Epidemiology* **172**, 478-487 (2010).
5. C. Yozgatligil, S. Aslan, C. Iyigun, I. Batmaz, Comparison of missing value imputation methods in time series: the case of Turkish meteorological data. *Theoretical and Applied Climatology* **112**, 143-167 (2013).
6. R. a. R. D. Little, Statistical Analysis with Missing Data. *Wiley, New York*, p. 381 (1987).
7. N. Bokde, M. W. Beck, F. Martínez Álvarez, K. Kulat, A novel imputation methodology for time series based on pattern sequence forecasting. *Pattern Recognition Letters* **116**, 88-96 (2018).
8. S. I. Khan, A. S. M. L. Hoque, SICE: an improved missing data imputation technique. *J Big Data* **7**, 37-37 (2020).
9. S. M. a. T. Bartz-Beielstein, imputeTS: Time Series Missing Value Imputation in R. *The R Journal* **9(1)**, 207–218 (2017).
10. Z. Che, S. Purushotham, K. Cho, D. Sontag, Y. Liu, Recurrent Neural Networks for Multivariate Time Series with Missing Values. *Scientific Reports* **8**, 6085 (2018).
11. W. Cao *et al.*, paper presented at the Proceedings of the 32nd International Conference on Neural Information Processing Systems, Montréal, Canada,  2018.
12. F. Biessmann *et al.*, DataWig: Missing Value Imputation for Tables. *J. Mach. Learn. Res.* **20**, 175:171-175:176 (2019).
13. T. Chen *et al.*, *MXNet: A Flexible and Efficient Machine Learning Library for Heterogeneous Distributed Systems*.  (2015).
14. V. Pedregosa F, Ga"el, Gramfort A, Michel V, Thirion B, Grisel O, et al., Scikit-learn: Machine learning in Python. *Journal of machine learning research* **12(Oct)**, 2825–2830 (2011).
15. T. H. J. Friedman, and R. Tibshirani., The elements of statistical learning. *Springer series in statistics Springer, Berlin,* **volume 1**,  (2001).
16. Y. Z. X. Yi, J. Zhang, and T. Li, St-mvl: filling missing values in geo-sensory time series data. *the Twenty-Fifth International Joint Conference on Artificial Intelligence, IJCAI 2016, New York, NY, USA, 9-15 July 2016, pages 2704–2710*,  (2016).